\documentclass{article}

\usepackage[numbers,sort]{natbib}

\usepackage[preprint]{neurips_2024}

\usepackage[utf8]{inputenc} 
\usepackage[T1]{fontenc}    
\usepackage{url}            
\usepackage{booktabs}       
\usepackage{amsfonts}       
\usepackage{nicefrac}       
\usepackage{microtype}      
\usepackage[dvipsnames,table,xcdraw]{xcolor}
\usepackage{nicematrix}
\usepackage{bbding}
\usepackage{floatrow}
\usepackage[font=small]{caption}
\usepackage{wrapfig}
\usepackage{hyperref}
\usepackage{graphicx}
\usepackage{multirow}
\usepackage{amsmath,amssymb,bm}
\usepackage{subcaption}
\usepackage{enumitem}
\usepackage[misc]{ifsym}
\usepackage{verbatim}
\usepackage{bm}
\usepackage{pifont}  
\usepackage{bbm}
\usepackage{esvect}
\newcommand{\cmark}{\ding{51}}%
\newcommand{\xmark}{\ding{55}}%

\hypersetup{
    colorlinks=true,
    linkcolor=red,
    citecolor=RoyalBlue,
    filecolor=magenta,      
    urlcolor=cyan,
}

\title{VCoME: Verbal Video Composition with Multimodal Editing Effects}

\begin{document}



\author{
    Weibo Gong, Xiaojie Jin$^{\dagger}$, Xin Li, Dongliang He, Xinglong Wu
    \\
    ByteDance Inc.
}

\def\thefootnote{\textdagger}\footnotetext{Corresponding author $<$\url{jinxiaojie@bytedance.com}$>$.}

\maketitle

\begin{abstract}
\label{abstract}
Verbal videos, featuring voice-overs or text overlays, provide valuable content but present significant challenges in composition, especially when incorporating editing effects to enhance clarity and visual appeal. In this paper, we introduce the novel task of \textit{verbal video composition with editing effects}. This task aims to generate coherent and visually appealing verbal videos by integrating multimodal editing effects across textual, visual, and audio categories. To achieve this, we curate a large-scale dataset of video effects compositions from publicly available sources. We then formulate this task as a generative problem, involving the identification of appropriate positions in the verbal content and the recommendation of editing effects for these positions. To address this task, we propose VCoME, a general framework that employs a large multimodal model to generate editing effects for video composition. Specifically, VCoME takes in the multimodal video context and autoregressively outputs where to apply effects within the verbal content and which effects are most appropriate for each position. VCoME also supports prompt-based control of composition density and style, providing substantial flexibility for diverse applications. Through extensive quantitative and qualitative evaluations, we clearly demonstrate the effectiveness of VCoME. A comprehensive user study shows that our method produces videos of professional quality while being 85\scalebox{1.25}{$\times$} more efficient than professional editors. To foster further research in this emerging direction, we will release all codes, models, and datasets at \url{https://github.com/LetsGoFir/VCoME}.

\end{abstract}

\section{Introduction}
\label{section:intro}

    Video platforms such as \textit{YouTube} and \textit{Reels} attract billions of users, establishing videos as the predominant medium for information dissemination. Among them, verbal videos are particularly valuable due to their rich content conveyed through voice-overs (as shown in \cite{maxwell2024voiceover}) or text overlays. However, composing verbal videos poses a significant challenge for individuals without professional expertise. The process involves using complex editing effects to highlight key information, enhance coherence, and increase visual appeal based on the multimodal content within the videos. For instance, key words are highlighted with text effects, emotions are intensified with sound effects, and video effects are used to increase attractiveness. These effects encompass numerous types and intricate combinations, making the composition a demanding task that requires a high level of editing skill and precision, which is difficult for non-professionals to achieve.
    
    
    This paper introduces the novel task of \textit{verbal video composition with editing effects} and presents a general framework, VCoME, to address it. The task is defined as follows: given a verbal video containing voice-over or text, the objective is to generate a composited video that incorporates multimodal editing effects across texts, audio and visual elements. The position and style of these effects should align with the video’s narrative to achieve coherence. By doing so, the composition can highlight key content in original videos, such as product advantages or key opinions, enhancing both clarity and aesthetic attractiveness.
    
    This task presents several challenges.
    First, there are no publicly available datasets for training and evaluation. It requires substantial effort to collect and annotate a large-scale video dataset. The diverse styles and content of verbal videos further complicate this process. Beyond the dataset issue, the inherent complexity of the task also poses significant challenges. High-quality video composition necessitates the seamless integration and harmony between effects and the associated multimodal content in videos, as well as coherence among the effects themselves. Given the vast number of effect types and combinations, the model must efficiently determine the appropriate effects. As the first exploration of this task, we need to address these challenges to deliver the optimal solution.
    
    We begin by constructing a verbal video with effects dataset sourced from publicly available videos on \textit{Jianying} \cite{capcutref}. Using video play-related metrics such as view counts and like counts, we curate a selection of high-quality videos with applied multimodal editing effects. Subsequently, we annotate the positions and types of effects from these videos to formulate the ground truth for this task, on which training and evaluation are performed.

    To tackle this task, we explore both classification-based and generative methods. Initially, we adopt a classification-based approach using transformers to directly output the IDs of recommended effects. Although straightforward, this approach encounters convergence issues, possibly due to its incapacity to model the task's high complexity. Motivated by the advanced reasoning and understanding capabilities of large multimodal models (LMMs), we transition to a generative approach. We first transform the task's outputs into a sequential composition target, enabling seamless integration with popular LMMs and end-to-end training. Next, we introduce a segment-level encoding method to extract embeddings for each input modality from video segments separated by sentences. This technique is crucial for the efficient fusion and understanding of correlated multimodal content. The resulting framework, VCoME, demonstrates strong capabilities to generate high-quality effect compositions from diverse multimodal content (visual, audio, and text) in input videos.

    In summary, this paper presents three key contributions:
    \begin{itemize}
        \item We introduce a novel task of verbal video composition with editing effects, which has wide applications in video editing. We curate a large-scale dataset of video effects composition to facilitate future research on this task.
        \item We formulate this task as a generative learning problem and propose a general multimodal framework VCoME that effectively models the relationship between multimodal inputs (visual, audio, and text) and various effects elements.
        \item Through both quantitative and qualitative analyses, we demonstrate that VCoME effectively acquires professional editing knowledge, producing high-quality video composition results. A comprehensive user study further verifies its effectiveness and efficiency.
    \end{itemize}


\section{Related work}

\begin{figure*}
        \includegraphics[width=\textwidth]{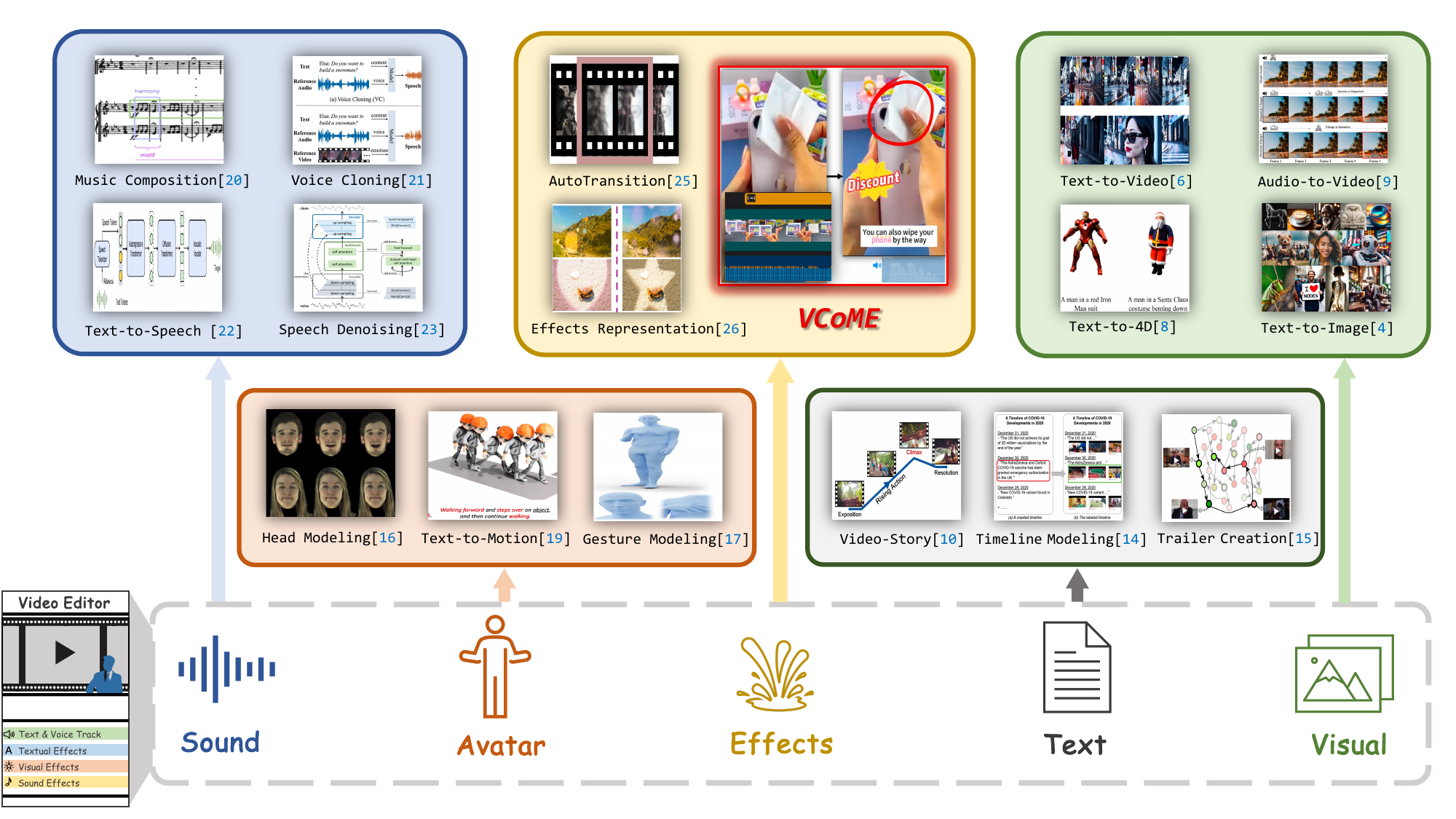}
        \caption{Tasks in video editing, grouped by target modality. Some tasks generate visuals, while others provide audio, text, and additional elements. Our method generates compositions with multimodal effects for verbal videos. Comprehensive video editing should consider the composition of all these elements.}
        \label{fig:tasks}
    \end{figure*}

    \paragraph{Video Editing}
    Video editing presents significant challenges for two primary reasons. First, it requires a deep understanding and manipulation of diverse elements across multiple modalities. Second, editing is highly subjective, heavily influenced by the creator's aesthetic preferences. 
    We categorize video editing tasks based on their objectives, as illustrated in Figure \ref{fig:tasks}. Visual generation models, including style transferring \cite{zhu2017unpaired}, text-to-image \cite{podell2023sdxl, shen2024many}, text-to-video \cite{sora} (mentioned by \cite{liu2024sora}), text-to-4D \cite{liang2024diffusion4d}, and audio-to-video \cite{jeong2023power}, are employed to generate visual materials. Furthermore, storytelling is crucial in video editing, which involves either narrative construction \cite{liu2024video, wang2019write, koorathota2021editing, yang2024synchronized} or timeline structuring \cite{choi2016video, papalampidi2023finding}. Additionally, avatar technology is valuable for verbal videos, allowing for control over head shapes \cite{hong2022headnerf} and human actions \cite{liu2023emage}, and can be driven by voice \cite{prajwal2020lip} or text \cite{guo2023momask}. Audio also plays a vital role in video editing, inspiring technologies such as AI music composition \cite{hernandez2022music}, voice cloning \cite{chen2022v2c}, speech processing \cite{anastassiou2024seed, kong2022speech}, and music-driven video montage \cite{liao2015audeosynth}. Integrating these multimodal elements into a cohesive video presents a significant challenge. In this work, we focus on verbal video composition with effects across multiple modalities.

    \paragraph{Effect Composition}
    Editing effects are essential for video production and come in various categories. Visual effects can enhance the visual experience and maintain viewer interest. For example, AutoTransition \cite{shen2022autotransition} recommends transition effects between video clips based on video and audio information, seamlessly connecting neighboring shots and scenes. Edit3k \cite{gu2024edit3k} learns the representation of more visual effects. Textual effects can significantly improve information delivery and the visual appeal of a video. MovieStudio \cite{CN116152393A} integrates both visual and textual effects, enabling it to generate music videos with a strong sense of rhythm like \cite{hyvine2024autocut}. Additionally, audio effects can strengthen emotional expression and atmosphere, providing audio cues and ambient sounds. Considering all these effects, VCoME introduces comprehensive video effects composition into the research domain for the first time, which is significant for both video editing applications and academic research.

    
    
    \paragraph{Large Multimodal Model (LMM)}
    Multimodal pre-training models \cite{li2023blip, bai2023qwen, achiam2023gpt, alayrac2022flamingo} demonstrate remarkable capabilities in understanding and generation across tasks such as video understanding, video captioning, and video-text retrieval. Video composition requires the comprehension of multimodal inputs and design knowledge, motivating us to utilize multimodal large models. We adopt large multimodal models like LLaVA \cite{liu2024visual} as our base framework. We use ViT \cite{dosovitskiy2020image} and ImageBind \cite{girdhar2023imagebind} to extract features from video and audio, respectively. These embeddings are then fed into large language models such as Mistral \cite{jiang2023mistral} or LLaMA \cite{touvron2023llama}. We treat the input text as temporal dependencies and sequentially predict the trigger positions and various editing effects, following the Chain-of-Thought \cite{wei2022chain} methodology.

\section{Task Definition}
\label{section:task_def}
    The objective of verbal video composition with effects is to determine both the positions and types of editing effects used in a verbal video. This aligns with the workflow of professional human creators with track editing tools, such as \textit{Adobe After Effects} (AE). Therefore, the verbal video composition process can be divided into two steps: positioning and recommendation. The former involves identifying key content within the video to enhance and trigger effects, while the latter entails generating appropriate effect names based on the position and input context.

    \begin{figure*}
        \includegraphics[width=\textwidth]{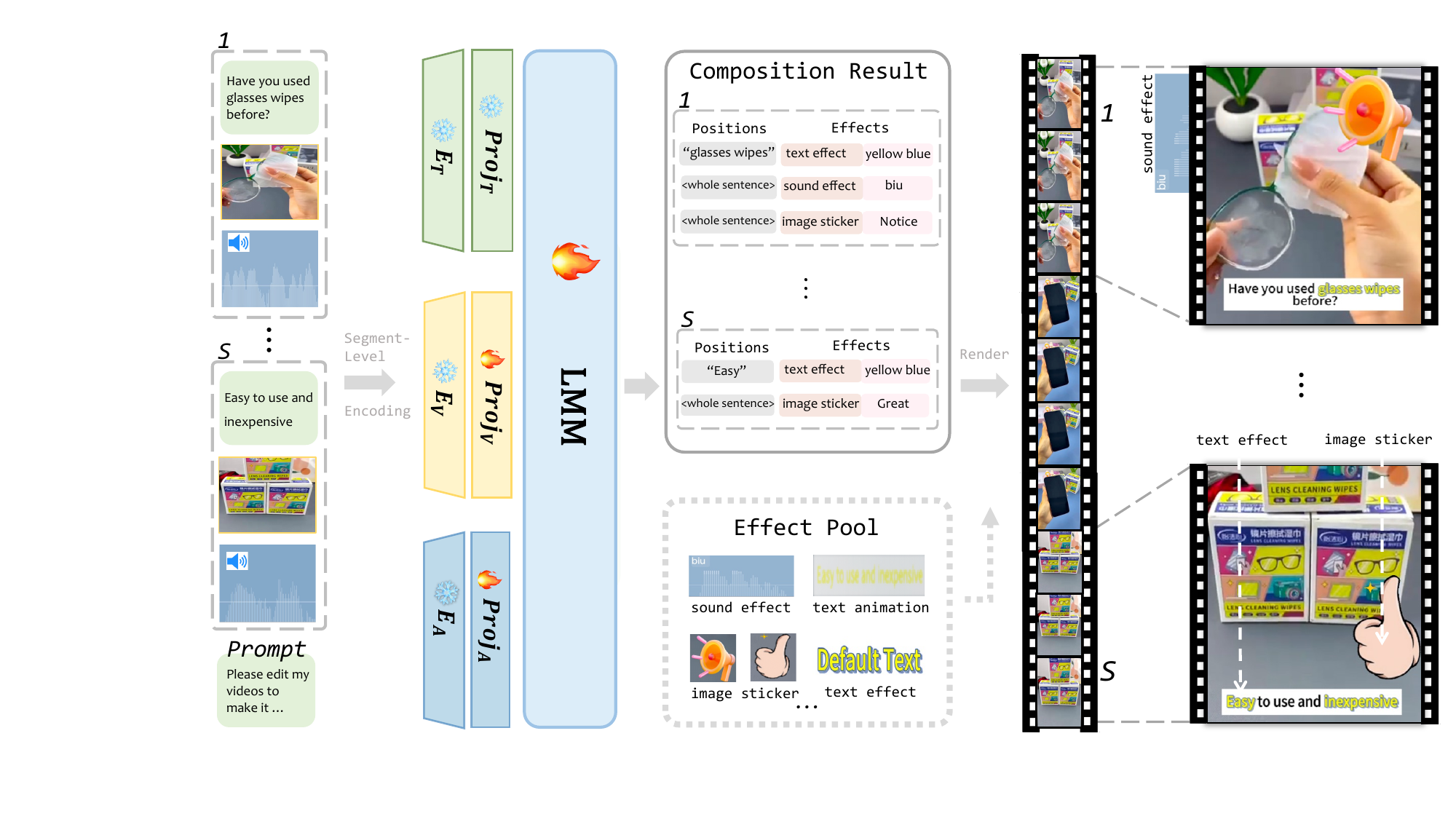}
        \caption{Our framework. The input data comprises three modalities and is segmented by sentences. After being processed by encoders and projectors, the multimodal tokens are uniformly handled by the LMM. Subsequently, the composition is generated, detailing the arrangement of various effects on the original video. The final video is rendered according to this composition, utilizing the effect pool.}
        \label{fig:pipeline}
    \end{figure*}
    
    \paragraph{Trigger positioning} The first step is to identify appropriate positions in the verbal video to apply (or trigger) editing effects, with the aim of highlighting the corresponding verbal content. These trigger positions are determined by one or several consecutive words within the verbal content. The applied effects start at the timestamp of the first word and end when the last word finishes. An example is shown in Figure \ref{fig:pipeline}, where we apply a text effect to highlight ``glass wipes'' along with a sound effect (named ``biu'' since sounds like it) and a image sticker (named ``Notice'' and looks like a horn) to all words indicated by ``<whole sentence>'' in segment 1, enhancing the presentation of objects in the video. This process improves the video's clarity and structure, making it better communicate the emphasized content and emotions to viewers.
    
    \paragraph{Effects recommendation} The second step involves determining which editing effects to apply to the video. This process requires ensuring stylistic consistency between the effects and their integration into the overall composition, while also considering the richness of the effects. Additionally, effects have controllable parameters such as intensity, transformation, and texture. To simplify the process, only effect types need to be predicted while setting the rest attributes of effects at default values. We carefully construct the effects pool, taking into account all three modalities, as illustrated in Figure \ref{fig:effects}. The categories include ``text animation'', ``text effect'' and ``text template'' for textual effects, ``image sticker'' for visual effects, and ``sound effect'' for audio effects. A detailed introduction to our effects can be found in Appendix \ref{effects_intro}.

    \paragraph{Notations} To carry out above two steps, we design the input and output format of VCoME as in Figure \ref{fig:pipeline}. Specifically, we introduce segment-level encoding, which cuts the input video into a series of segments $X^{\text{content}} = \{ x_1, {\dots}, x_{S}\}$ based on the sentence breaks in verbal content, where $S$ is the number of video segments. Then for the $i$-th segment, we have $x_{i} = \{s_i, v_i, a_i\}$ as input, in which $s_i$, $v_i$, $a_i$ represents the sentence, video clip and audio clip, respectively. This division is performed within a single video. When multiple videos are provided, we first combine them into a single video and then apply the same procedure.
    
    For the output, we arrange the recommended effects according to the segments indices. The learning objective for the $i$-th segment includes multiple editing effects, denoted as
    \small{
    \begin{equation*}
    \label{ground_truth_eq}
    y_{i} = \big(i, (l_1, e_1),(l_2,e_2), \dots\big).
    \end{equation*}
    }\normalsize
    \noindent
    Besides the index $i$, each element in $y_i$ is represented in the format of $(l, e)$. $e$ is the name of effect and $l$ is its trigger position. After processing $X^{\text{content}}$ through multimodal encoders, the context information is mapped to a unified joint space. Subsequently, the LMM processes this information and autoregressively generates the composition text, which is the concatenation of $y_i$ in the form $Y = (y_1, \dots, y_S)$.
    The display duration of each effect aligns with its corresponding trigger position within the segment.
    

\section{Video Composition Dataset}
\label{section:dataset}

\subsection{Data Collection}

    A vast number of high-quality publicly available videos can be found on popular video sharing platforms. These videos are created by a large amount of creators, displaying a wide range of design styles and narrative content. Well-produced verbal videos often adopt structured storytelling and incorporate various editing effects, such as image stickers, text/video effects, and sound effects. To construct our dataset, we collect videos that contain both verbal content and editing effects, and apply annotations.
    The parameters of effects are diverse and intricate. For example, text animations have intensity and speed parameters, while sound effects have volume and duration parameters. To focus on the core positioning and recommendation of effects, we do not infer the specific parameter values. Instead, we directly use the default parameters of effects for rendering purposes. This simplification facilitates the learning of video composition while preserving its integrity by retaining the key attributes of the effects.


    \subsection{Data Filtering}
    \label{data_filtering}
\begin{figure}
    \centering
        \begin{minipage}{0.33\textwidth}
          \centering
          \includegraphics[width=\textwidth]{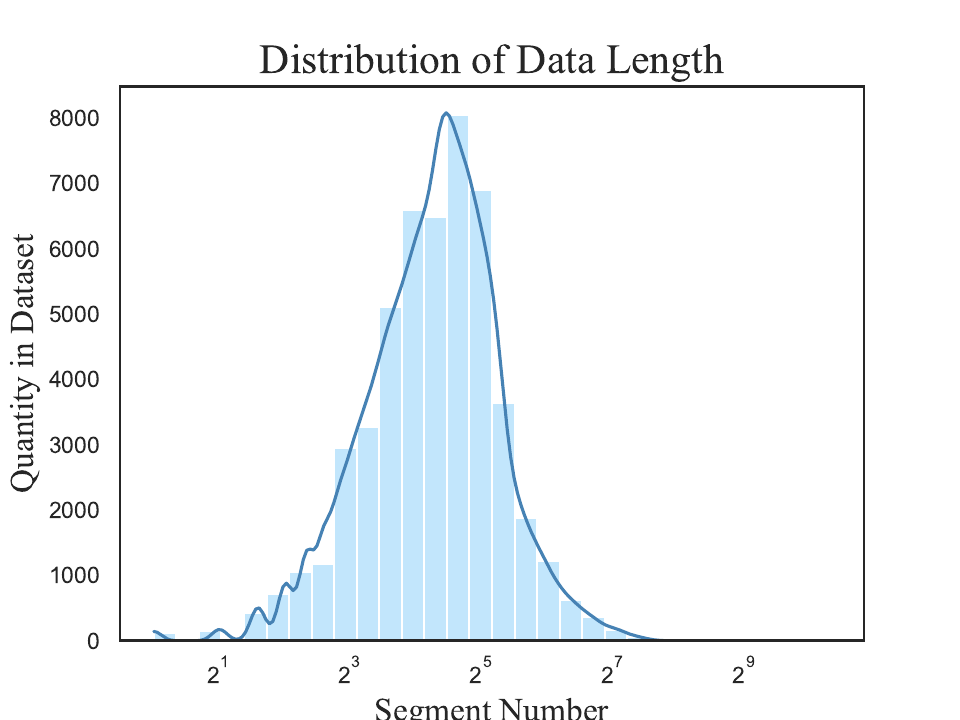}
          \subcaption{Distribution of data length.}
          \label{fig:data_length}
        \end{minipage}\hfill
        \begin{minipage}{0.33\textwidth}
          \centering
          \includegraphics[width=\textwidth]{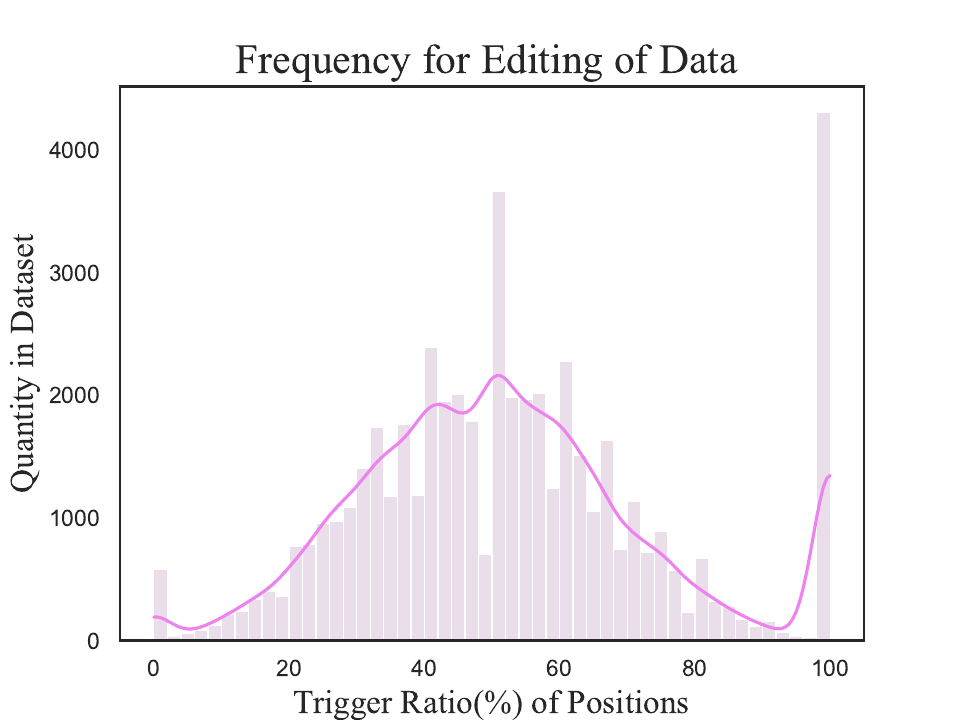}
          \subcaption{Distribution of trigger positions.}
          \label{fig:decoration_ratio}
        \end{minipage}\hfill
        \begin{minipage}{0.33\textwidth}
          \centering
          \includegraphics[width=\textwidth]{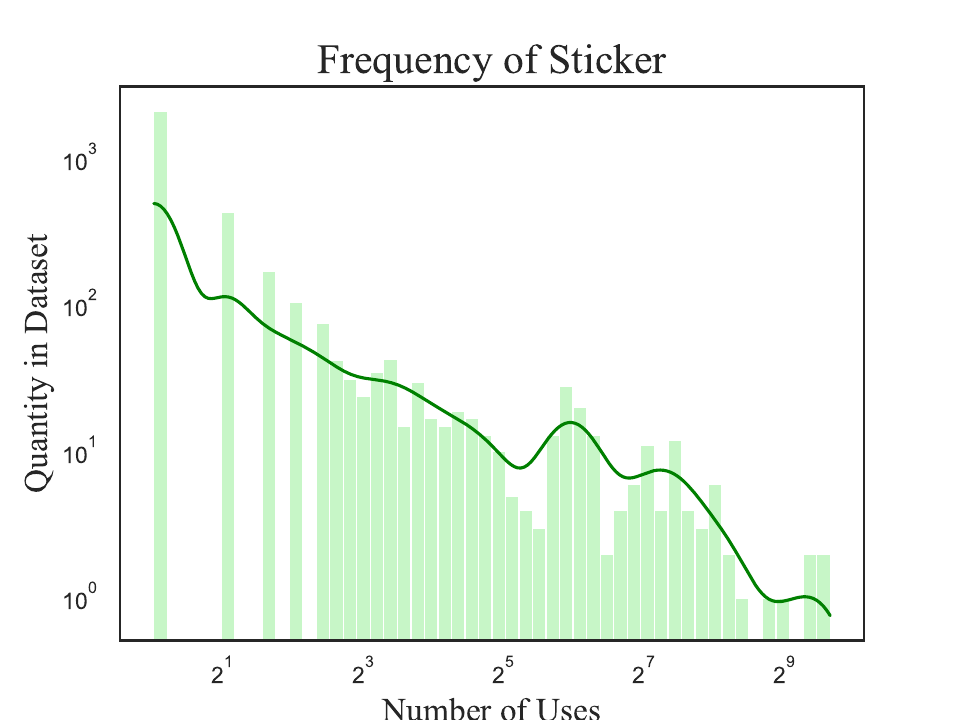}
          \subcaption{Distribution of stickers used.}
          \label{fig:sticker_distribution}
        \end{minipage}
    \caption{Distribution of our dataset.}
    \end{figure}

    The distribution of video length exhibits significant variation, as depicted in Figure \ref{fig:data_length}. Some samples consist of only a few sentences, while some exceed the maximum length of our model, extending to over a hundred sentences. To address this issue, we employ two strategies: filtering out both data with fewer than three sentences and truncating excessively long texts. The second strategy allows us to utilize a sufficient amount of data within the model's length limitations and acts as a form of data augmentation since the truncated data does not begin with the first segment. Additionally, the frequency of trigger positions (Figure \ref{fig:decoration_ratio}) and the usage distribution of effects, such as image stickers (Figure \ref{fig:sticker_distribution}), are imbalanced. This phenomenon increases the difficulty of this task. Finally, we gather a dataset comprising approximately 53,000 instances, which we subsequently divide into 51000 samples for the training set and 2,100 samples for the validation set. Each data sample consists of multiple segments.
    
\section{Verbal Video Composition}
\label{method_composition}

\subsection{Context Modeling} \label{section:model structure}
    We formulate the task as a multimodal generation problem. The primary challenge we address is enabling the LMM to comprehend the relationship between multimodal inputs and the composition of effects. Given that visual and tonal changes within each sentence are minimal, we design segment-level encoding to achieve efficient understanding and multimodal fusion as in Figure \ref{fig:pipeline}. This process involves segmenting the video at the sentence level, followed by encoding in three different modalities.

    To process visual and text input, we adopt a similar approach to LLaVA~\cite{liu2024visual}. We pre-train the model on image-text pair data, developing a projector from the visual encoder to the large multimodal model. For finetuning on the video composition dataset, we simplify the process by extracting a single frame for each segment \(x_i\), as the frames within a single segment are similar, and integrating it into the LMM for comprehension. Specifically, we obtain visual embeddings using $\texttt{E}_V$, the pre-trained Vision Transformer in CLIP \citep{radford2021learning} and map them to the latent space of the large multimodal model using an aligned MLP $\texttt{Proj}_V$. Because the input can be potentially lengthy, e.g., more than 100 segments, it is not feasible to retain all ViT \citep{dosovitskiy2020image} patches for each frame. Therefore, we preserve only the [CLS] token for each frame. Similarly, we process audio input with the encoder-projection structure. In our experiments, we extract audio features using ImageBind \citep{girdhar2023imagebind} as $\texttt{E}_A$ for each segment.
    
    It is noteworthy that encoders for both visual and audio inputs can be trivially replaced by other widely used modality encoders. We project the visual and audio inputs into the same dimension through independent linear transformations $\texttt{Proj}_V$ and $\texttt{Proj}_A$. The embeddings of sentences, video clips, and audio clips can be represented as follows:
    \[
    \begin{array}{ccc}
    f_s = \texttt{Emb}(\texttt{tokenizer}(s_i)), &
    f_v = \texttt{Proj}_V(\texttt{E}_V(v_i), &
    f_a = \texttt{Proj}_A(\texttt{E}_A(a_i)).
    \end{array}
    \]
    \noindent
    $s_i$, $v_i$ and $a_i$ represent the sentence, video clip, and audio clip, respectively. $f_s$, $f_v$, and $f_a$ represent their corresponding embeddings. The terms $\texttt{Emb}$ and $\texttt{tokenizer}$ correspond to the $\texttt{Proj}_T$ and $\texttt{E}_T$ in Figure \ref{fig:pipeline} respectively.
    Subsequently, we input these embeddings into the LMM, along with learnable positional embeddings, enabling the transformer to effectively capture segment-level information. The model then sequentially generates the composition results.

    \subsection{Composition Learning}
    \label{section:compostion_learning}
    As introduced in Section \ref{section:model structure}, the multimodal embeddings are extracted using their respective encoders and serve as contextual information. During the composition process, we use learnable linear transformations to map the video and audio embeddings into the large multimodal model, facilitating the positioning and recommendation of effects.
    
    We denote 
    \small{
    \begin{align*}
        E^{\text{elem}}= \{& e^{\text{text-animation}}_1, e^{\text{text-effect}}_1, e^{\text{text-template}}_1, 
                e^{\text{sound-effect}}_1, e^{\text{image-sticker}}_1, \dots, \notag\\
                & e^{\text{text-animation}}_{N_{\text{t-ani}}}, 
                e^{\text{text-effect}}_{N_{\text{t-eff}}}, e^{\text{text-template}}_{N_{\text{t-tem}}},
                e^{\text{sound-effect}}_{N_{\text{s-eff}}}, e^{\text{image-sticker}}_{N_{\text{i-stk}}}\},
    \end{align*}
    }\normalsize
    \noindent
    as the set of editing effects where $N_{\text{t-ani}}$, $N_{\text{t-eff}}$, $N_{\text{t-tem}}$, $N_{\text{s-eff}}$, $N_{\text{i-stk}}$ are numbers of each category of effects. $N_{\text{i-stk}}=10,000$ and others are set as 1,000 in experiments. 
    The input is \(X = \{X^{\text{content}}, X^{\text{prompt}}\}\), where \(X^{\text{content}}\) represents the multimodal input in a sample, and \(X^{\text{prompt}}\) represents the prompt, which is discussed in Section \ref{section:prompt_design}. For each segment \(x_i\) in \(X^{\text{content}}\), we generate the corresponding composition result \(y_i\) using the multimodal large model \({p_\theta}(y_i|X, y_{j<i})\). The target is formed as \(Y = (y_1, \dots, y_S)\). A detailed introduction and design concepts for our format are presented in Appendix \ref{learning_objects_introduction}.


    \subsection{Prompt Design}
    \label{section:prompt_design}
    Inspired by \cite{white2023prompt}, we design prompts to cater to diverse user intents. For instance, users may control the frequency of effect occurrence or choose specific categories of effects. We design several prompts to address this issue. For instance, during training, we use prompts such as ``Please edit a video with a suitable frequency of trigger positions and include image stickers'' and ``Please edit a video with a 50\% frequency of trigger positions and make use of animated text.'' During inference, prompts like ``Please edit a video with a 70\% frequency of trigger positions, simultaneously incorporating image stickers and animated text'' can be used to generate the desired result. This approach leverages the capabilities of large multimodal models to generate more diverse data, effectively addressing the long-tail phenomenon and facilitating intelligent interaction for users. As shown in Figure \ref{fig:sticker_distribution_prompt}, we get more image stickers when prompting with ``more image stickers'' compared to \ref{fig:sticker_distribution_test} without prompts.

    By considering all input \( X = \{X^\text{content}, X^\text{prompt}\} \), we can offer users more diverse and enriched compositions through user-friendly interactions.

    \subsection{Training and Evaluation}
    Following LLaVA, we adopt a two-stage training procedure: pre-training and finetuning. For evaluation, we establish both objective and subjective metrics specific to this task. The details of our training and evaluation process are outlined below.

    \paragraph{Training}
    We employ the cross-entropy loss as our model's learning objective:
    \small{
    \begin{equation}
    \label{eq:final-loss}
    \mathcal{L}(X) = \prod_{i=1}^{S}{p_\theta\left({y}_i| y_{j<i}, X^{\text{content}}, X^{\text{prompt}} \right)},
    \end{equation}
    }\normalsize
    \noindent
    where \(\theta\) represents the parameters of our model and \(S\) is the number of segments. By optimizing Equation \ref{eq:final-loss}, the model learns to incorporate the multimodal context, prompt, and previous composition results. This optimization encourages the identification of suitable trigger positions and the recommendation and combination of appropriate effects during the generation process.
    
    \smallskip
    
    \paragraph{Evaluation}
    We define the evaluation metrics across two dimensions: objective and subjective. The objective metrics primarily focus on the accuracy of trigger positions and corresponding effects. We use the Dice similarity \cite{zou2004statistical} to calculate the average accuracy of trigger positions, as shown in Equation \ref{eq:metric_word_precision}. Once a trigger word is predicted correctly, we calculate the precision of the effect category using Equation \ref{eq:metric_type_precision}. In addition to words, effects like text animations, sound effects, or image stickers are applied to all words in one sentence. Thus, for these sentence-level effects, we calculate the recall of the corresponding elements within the specific segment, as shown in Equation \ref{eq:metric_sentence_recall}.

    \small{
    \begin{align}
        M^\text{word\_accuracy}_X &= \frac{1}{|Y_w \cup Y^{\prime}_w|} \sum_{i \in |Y_w \cup Y^{\prime}_w|} \text{Dice}\left(w_i, w^{\prime}_i\right) \label{eq:metric_word_precision} \\
        M^\text{elem@word}_X &= \frac{1}{|Y_w|} \sum_{i \in |Y_w|} 
            \begin{cases}
                1, & \text{if Dice}(w_i, w^{\prime}_i) \geq 0.5 \text{ and Eq} (w_i, w^{\prime}_i)\\
                0, & \text{otherwise}
            \end{cases} \label{eq:metric_type_precision} \\
        M^\text{elem@sentence}_X &= \frac{1}{|Y_h|} \sum_{k \in |Y_h|} \frac{{|h_k \cap h^{\prime}_k|}}{{|h_k \cup h^{\prime}_k|}}.\label{eq:metric_sentence_recall}
    \end{align}
    }\normalsize
    \noindent \( Y = \{Y_w, Y_h\} \) is the ground truth, which includes elements with trigger positions \( l \) and effects \( e \) as shown in Section \ref{ground_truth_eq}. \( Y_w \) comprises elements corresponding to words, denoted by \( w_i \), while \( Y_h \) contains elements for the entire sentence, denoted by \( h_k \). \( Y^{\prime} \) represents the results generated by the model, with predicted elements denoted as \( w^{\prime}_i \) and \( h^{\prime}_k \), respectively. The metric \( M^\text{word\_accuracy}_X \) represents the word accuracy of data sample \( X \), where ``Dice()'' calculates the Dice similarity of \( l \) for the ground truth \( w_i \) and predicted elements \( w^{\prime}_i \). When calculating \( M^\text{elem@word} \) in Equation \ref{eq:metric_type_precision} for \( w^{\prime}_i \) and \( w_i \), we set the value to 1 if the matched word achieves a Dice similarity greater than 0.5 and the effects category is correct, denoted by ``Eq\( (w_i, w^{\prime}_i) \)''; otherwise, it is set to 0. For the convenience of comparing competing methods, we aggregate these three indicators into an overall score.

    Subjective evaluation involves using ``Win/Tie/Loss'' and ``Mean Opinion Score'' to assess various qualitative aspects. In consultation with professionals, we establish evaluation dimensions including video aesthetics and decorative richness, along with their corresponding scores. For each sample in the user study, we conduct blind testing with volunteers to obtain ratings.

\section{Experiments}

    \subsection{Implementation Details} \label{section:exp}
        
        \paragraph{Model details}
        We utilize the open-source bilingual Chinese-LLaMA-2 \cite{cui2023efficient} as the backbone, incorporating CLIP-ViT-L/14 Vision Transformer for visual feature extraction. For audio, we employ ImageBind \cite{girdhar2023imagebind} to extract local audio features. Both modalities are linearly projected to a shared latent space and used as input embeddings for the LMM. The LMM supports 2048 tokens and comprises approximately 40 stacked transformer blocks. To save GPU memory, we freeze the parameters of the Vision Transformer and train only the LMM and the projectors for each modality.
        
        \smallskip

        \paragraph{Data pre-processing and training}
        For each video clip in a data sample, we extract the middle frame as the visual input. These input frames have a resolution of \(\text{336}\times \text{336}\) and are represented by a [CLS] token with a dimension of 2048 as the visual feature. ImageBind generates 1024-dimensional feature vectors, sampled every 0.5 seconds, but limited to a maximum of three embeddings per video clip due to memory constraints. In all experiments, we employ the Adam optimizer with an initial learning rate of 1e-4, decayed using the cosine strategy. During pre-training, we align the visual model with the large language model using the LLaVA pipeline and align the audio features to the joint space using a projector on our dataset. The batch size for each device is 16. Our experiments are conducted using 8 NVIDIA A100 GPUs, with each training session taking approximately 4.5 hours.

        \smallskip
        
        \paragraph{Metrics}
        In Section \ref{section:compostion_learning}, we define three objective metrics to evaluate the performance of the composition model: word accuracy, effect precision for words, and effect recall for sentences. The overall score is calculated as the sum of these individual metrics. Due to space limit, the variations in experimental results of our best-performing method are provided in Appendix \ref{section:error bar}.
        
        In addition to the objective metrics, we consider many other factors when performing subjective evaluation. For further details, please refer to our user study Section \ref{user_study}.

        \smallskip

        \paragraph{Baselines}
        \label{baselines}
        The list of compared baseline methods is shown in Table \ref{tab:main_result}. Initially, we implemented a two-stage baseline, which utilizes BERT for trigger positioning, followed by a classifier for effect recommendation. Despite its simplicity and efficiency, this method does not converge during the classifier's training phase. Next, we employ a Large Multimodal Model (LMM) and successfully generate reasonable results. We then design the second baseline, which involves learning the indices within the sentence as trigger positions. This means that the trigger position \( l \) in \( y_i \) is represented by indices, such as ``4-5'' for ``glass wipes'' in Figure \ref{fig:pipeline}, rather than the words themselves.
        This approach reduces the output token length of the large multimodal model compared to direct word prediction. Lastly, we explore the use of prompts to increase the frequency of trigger positions, reducing empty predictions for each segment and applying more effects. As shown in Figure \ref{fig:decoration_ratio_prompt}, we observe a higher ratio of trigger positions with prompting more effects, compared to Figure \ref{fig:decoration_ratio_test} without prompts, thereby achieving better alignment with the ground truth.
        
        \setlength{\tabcolsep}{4pt}
        \begin{table}[tbp]
        \centering
        \begin{center}
            \begin{tabular}{c|ccc|c}
                \toprule[2pt]
                Method & Word Accuracy & Elem@Word & Elem@Sentence & Overall Score $\uparrow$ \\
                \midrule
                $\text{Baseline}_\text{two-stage}$  & 13.78\% &  0  &  0  & 13.78 \\
                
                $\text{Baseline}_\text{trigger-index}$ & 32.05\% & 65.31\% & 28.5\% & 125.86 \\
                \midrule
                Ours & 34.46\% & \textbf{69.08\%} & 30.52\% & 134.06 \\
                
                Ours\textsubscript{prompt} & \textbf{37.88\%} & 68.25\% & \textbf{31.90\%} & \textbf{138.03} \\
                \bottomrule[2pt]
            \end{tabular}
            \caption{Main results. The baseline of ``two stage'' represents the BERT followed by a classifier, and the baseline of ``trigger index'' predict the indices within strings rather than words. See more details in Section \ref{baselines}.}
            \label{tab:main_result}
        \end{center}
        \end{table}
        \setlength{\tabcolsep}{1.4pt}
        
    \subsection{Ablation Studies and Comparisons} \label{section:ablations}
        
        We first compare inputs across different modalities to highlight the advantages of multimodal context. Then, we perform comparisons involving varying data scales and model sizes. Finally, we explore different learning targets to validate the superiority of our target design.
        
        \smallskip
        \paragraph{Multimodality and multilinguality} 
        In this experiment, we examine the influence of the base model and multimodal inputs on the quality of the generated outputs, as illustrated in Table \ref{tab:model_and_modal}. The term ``bilingual'' denotes the use of a bilingual pre-trained model. Utilizing English-only models, such as LLaVA-vicuna, as the base model results in suboptimal generation outcomes, primarily due to their limited ability to handle multiple languages. Comparing the second and third rows of the table, we observe that the absence of the visual modality significantly impairs the model's performance, underscoring the importance of visual content for our task. Finally, the last row demonstrates that incorporating all modal inputs yields the greatest enhancement in the quality of the generated compositions.
        
        \begin{table}[t]
        \centering
        \small
            \begin{subtable}[h]{0.48\textwidth}
                \centering
                \setlength{\tabcolsep}{5pt}
            \begin{tabular}{cc|c|c}
                \toprule[2pt]
                \multicolumn{2}{c|}{Modal} & \multirow{2}{*}{Bilingual} & \multirow{2}{*}{Overall Score $\uparrow$} \\
                \cline{1-2}
                
                Visual & Audio & & \\
                \midrule
                \cmark & \textcolor{lightgray}{\xmark} & \textcolor{lightgray}{\xmark}& 111.92 \\
                \textcolor{lightgray}{\xmark}& \textcolor{lightgray}{\xmark}& \cmark & 128.83 \\
                \cmark & \textcolor{lightgray}{\xmark}& \cmark & 134.07 \\
                \cmark & \cmark & \cmark & \textbf{138.03} \\
                \bottomrule[2pt]
            \end{tabular}
                \caption{The impact of different modality inputs on the results and the effectiveness of bilingual pre-trained models.}
                \label{tab:model_and_modal}
            \end{subtable}
            \hfill
            \begin{subtable}[h]{0.48\textwidth}
                \centering
                \setlength{\tabcolsep}{5pt}                
                \begin{tabular}{c|c|c}
                    \toprule[2pt]
                    Order & Indices & Overall Score $\uparrow$ \\
                    
                    \midrule
                    $\text{Order}_{\text{random}}$      & \cmark     & 116.49 \\
                    $\text{Order}_{\text{string}}$      & \cmark     & 122.35 \\
                    $\text{Order}_{\text{category}}$        & \cmark     & 123.21 \\
                    \midrule
                    $\text{Order}_{\text{time}}$        & \textcolor{lightgray}{\xmark}    & 114.66 \\
                    $\text{Order}_{\text{time}}$        & \cmark     & \textbf{130.73} \\
                    \bottomrule[2pt]
                \end{tabular}
                    \caption{The impact of target design on the generated results, including the order of effects in target and the indices of each segment.}
                    \label{tab:target_design}
            \end{subtable}
            \caption{Ablations on model design and target design. We find both the multimodal input and ordered targets are important for learning.}
        \end{table}

        \smallskip
        \paragraph{Scale} 
        In the data scale experiment, we constructed training sets by randomly selecting 10,000, 20,000, 50,000, and 300,000 instances, respectively. Audio input is not included for efficiency. As presented in Table \ref{tab:scale_exp}, our models show consistent improvement in performance with increased data scale.  This indicates the strong potential of our model to further enhance its performance with additional data.  

        \setlength{\tabcolsep}{4pt}
        \begin{table}[tbp]
        \centering
        \begin{center}
            \begin{tabular}{c|c|ccc|c}
                \toprule[2pt]
                Data Scale & Model Scale & Word Accuracy & Elem@Word  & Elem@Sentence & Overall Score $\uparrow$ \\
                \midrule
                50k  & 7B  & 34.78\% & 68.50\% & 27.45\% & 130.73 \\
                
                50k  & 13B & 39.33\% & 69.62\% & 27.90\% & 136.85 \\
                
                \midrule
                10k  & 7B  & 24.55\% & 55.17\% & 16.93\% & 96.65  \\
                20k  & 7B  & 28.66\% & 61.62\% & 21.99\% & 112.27 \\
                300k & 7B  & \textbf{45.49\%} & \textbf{77.70\%}  & \textbf{42.15\%} & \textbf{165.34} \\
                \bottomrule[2pt]
            \end{tabular}
            \caption{The ablation results of different data scales and model scales. The model performance increases as the data and model sizes scale up. }
            \label{tab:scale_exp}
        \end{center}
        \end{table}
        \setlength{\tabcolsep}{1.4pt}

        Furthermore, in the ablation on model size, we utilize a larger 13B language model, which yields better generation results compared to the 7B model.

        \smallskip
        \paragraph{Target design} 
        In this experiment, we investigate the impact of different learning objectives, as shown in Table \ref{tab:target_design}. Audio is excluded from this experiment to expedite the training process. The results in the first column demonstrate that designing a more structured learning objective for each segment enhances the quality of the model's generation. This approach aligns with the workflow of human designers, who typically apply effects to their videos in chronological order, resulting in a more logical and organized composition. Specifically, ``random'' represents no specific order for the targets within each segment, while ``category'' indicates that the effects are ordered by category, such as predicting ``text-effect'' first, followed by ``sound-effect.'' ``String'' refers to the order of trigger words in the sentence, and ``time'' pertains to the timestamps in the composition information. The second column shows that the performance is significantly impaired when the indices of the segments are not explicitly indicated in $X^\text{content}$ and $Y$.

    \subsection{User Study}
    \label{user_study}
\begin{figure}
            \centering
            \includegraphics[width=\textwidth]{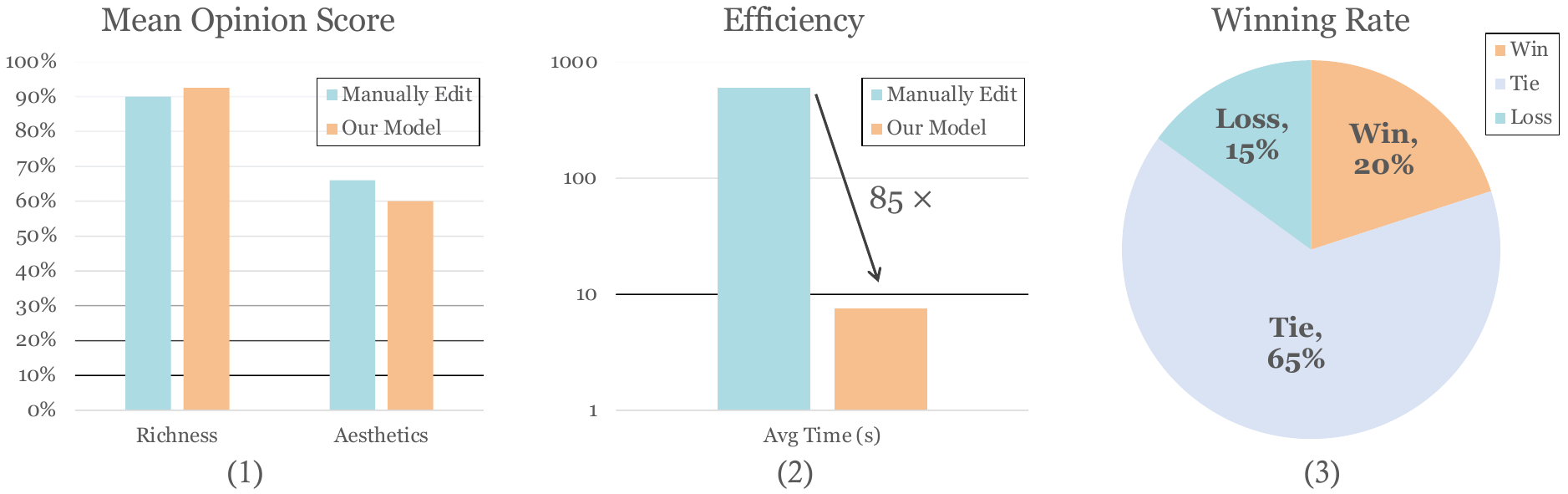}
            \caption{User study. (1) We conduct a comparison using mean opinion scores, assessing both content richness and visual aesthetics. (2) The time required for manual editing versus model inference. (3) The comparative analysis reveals that the acceptability of both the model-generated and manually edited videos is similar.}
            \label{fig:user_study}
        \end{figure}

        Due to the artistic and subjective nature of benchmarking video composition, user studies are of utmost importance. We collect videos from public sources covering various domains such as food, parenting, real estate, and broad marketing. These videos are then manually edited by professional video editors to compare with our model's composition results. We conduct a blind test, enlisting volunteers to assess the scores and determine the win/tie/loss outcomes for each result pair. The evaluation results are detailed in Figure \ref{fig:user_study}. Our model demonstrates performance comparable to that of human creators. We believe that further enhancing the data quality will lead to even more favorable outcomes. Notably, our model significantly improves editing efficiency, reducing the editing time for verbal videos by approximately 85 times.

\section{Limitations}
\label{section:limit}
    Our method has the following limitations.
    First, we do not account for all categories of effects, such as facial effects, video effects, and illustrations, necessitating the collection of more extensive data to encompass these categories. Additionally, the granularity of our visual encoding method is relatively coarse to save memory, extracting only one frame per segment. This approach may result in the loss of visual details. Efficient visual representations such as dynamic image/video tokenization may help achieve better trade-off between performance and efficiency.
    Moreover, it is essential to introduce multimodal generation techniques to directly create editing effects, auxiliary shots, and even all the targets in Figure \ref{fig:tasks} with a unified model. 
    
\section{Conclusion}
\label{section:conclusion}
    The rapid development of social media platforms and editing tools has created a significant demand for video editing, highlighting the need for more efficient methods to reduce the learning curve and usage cost. In this paper, we propose a novel task of verbal video composition with effects and present a general framework. We build the first multimodal video composition dataset, formulate the task as a generation problem, and solve it by leveraging the understanding and reasoning capabilities of large multimodal models (LMMs). Our approach proves effective in generating high-quality video compositions. We hope our work will benefit non-professionals and inspire further research on video editing tasks. Future work includes expanding to more video editing elements, such as music, facial effects, and virtual humans, and supporting more user intentions to control the density, themes, and parameters of effects.

\bibliographystyle{unsrtnat}
\bibliography{Ref}
\label{reference}







\newpage
\appendix

\section{Appendix / supplemental material}

    \subsection{Effects Introduction}
    \label{effects_intro}
    We have effects in 3 modalities, as follows:
    \begin{itemize}
    \item[1] Textual effects: This category includes text effects, text animations, and text templates. Text effects involve modifying attributes such as color, style, and other features. Text animations incorporate transformations like rotation and translation applied to the text. Text templates offer more intricate designs and include supplementary graphic elements. These effects are categorized into approximately 1,000 distinct types.

    \item[2] Audio Effects: We focus on sound effects for audio, such as ``ding'' or ``whoosh,'' which are triggered based on the emotional tone and content of the text. These effects capture viewers' attention and effectively enhance the video's clarity. Although music is also available, we do not focus on it in this context. There are approximately 1,000 distinct sound effects available in total.
    
    \item[3] Visual effects: Stickers. We emphasize stickers due to their vast types and rich themes. These include small images depicting a range of expressions. Recommending appropriate stickers is a challenging task due to their extensive variety. Our dataset comprises approximately 10,000 distinct stickers.
    \end{itemize}

    \subsection{Learning Objects}
    \label{learning_objects_introduction}
    The following part provides a comprehensive introduction to the input and learning objectives designed by us.
    \begin{itemize}
    
        \item \textbf{Context $X^{\text{content}}$} This part describes the multimodal information provided, which includes video clips, audios, and texts. In our dataset, there are instances where the video and audio inputs are empty. The collected data consists of verbal videos, typically containing around 20 sentences and hundreds of corresponding tokens.
        
        \item \textbf{Prompt $X^{\text{prompt}}$} In real-world application scenarios, users may desire control over decoration effects, including the frequency and specific categoryto be used. This can result in distributions that differ from the training data. To assess the model's responsiveness to user intent, we design task prompts that encompass these two scenarios.
        
        \item \textbf{Index $i$} The indexing of segments is designed to assist the model in effectively identifying contextual information within the given context, thus capturing the content of the response. Additionally, it enables a more intuitive observation and interpretation of the model's output.

        \item \textbf{Trigger position $l$} In our task, each sentence contains trigger positions represented by specific words known as trigger words, which are derived from the original text. By incorporating the learning objective of inferring trigger words, we enable the model to learn the editing workflow in a more organized manner. For instance, in a sentence like ``The cream bread is delicious'', the model identifies ``delicious'' as a trigger word, indicating the need for editing, and then determines the appropriate effect to apply. In addition to trigger words, we can also employ trigger captions, trigger emotions, or trigger intervals to leverage the multimodal input and incorporate reasoning for editing.
        
        \item \textbf{Editing Effects $e$} Once the trigger position is identified, the model can infer the specific type of effect without requiring additional tokens in the vocabulary. We use the ``category-name'' representation to uniquely denote each effect. This approach leverages the LMM's ability to effectively explore the relationship between effect names and video context. Furthermore, adding extra tokens to the vocabulary can complicate the initialization process. For instance, random initialization could result in the loss of valuable information of effects.
    \end{itemize}

    \begin{figure}
        \centering
        \includegraphics[width=\textwidth]{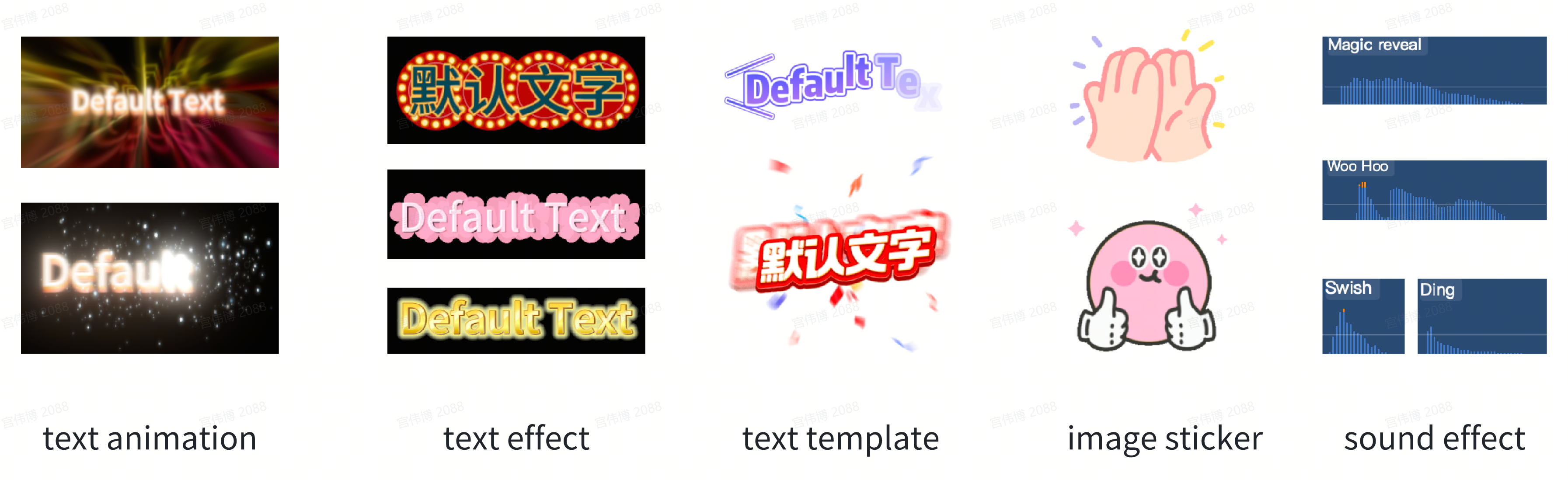}
        \caption{Categories of editing effects used in our task.}
        \label{fig:effects}
    \end{figure}

    \begin{figure}
    \centering
        \begin{minipage}{0.45\textwidth}
          \centering
          \includegraphics[width=\textwidth]{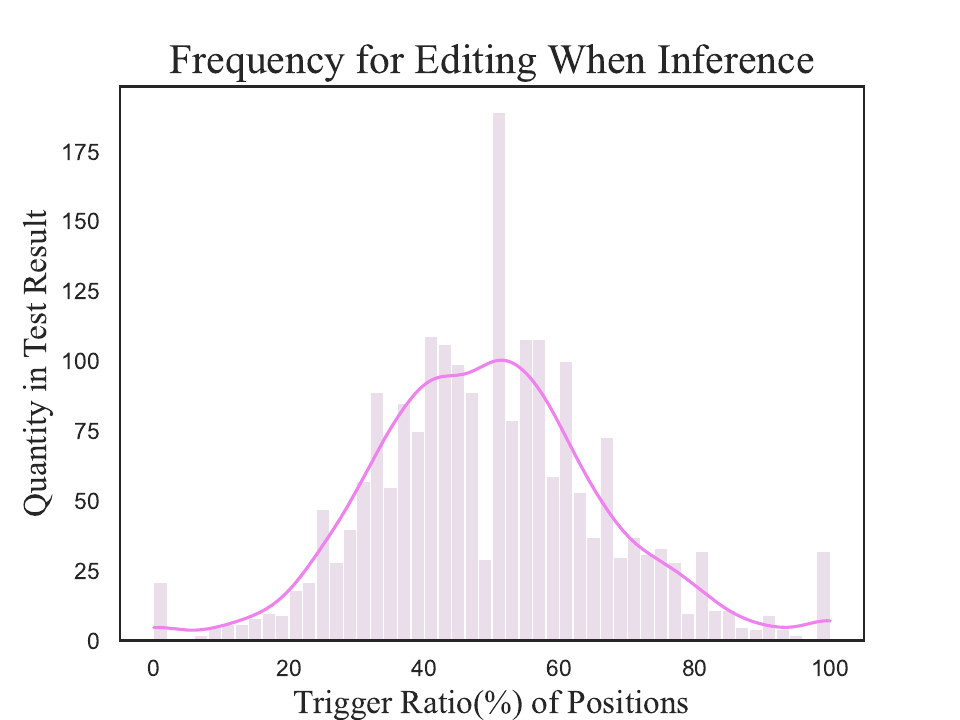}
          \subcaption{Distribution of trigger positions in test.}
          \label{fig:decoration_ratio_test}
        \end{minipage}\hfill
        \begin{minipage}{0.45\textwidth}
          \centering
          \includegraphics[width=\textwidth]{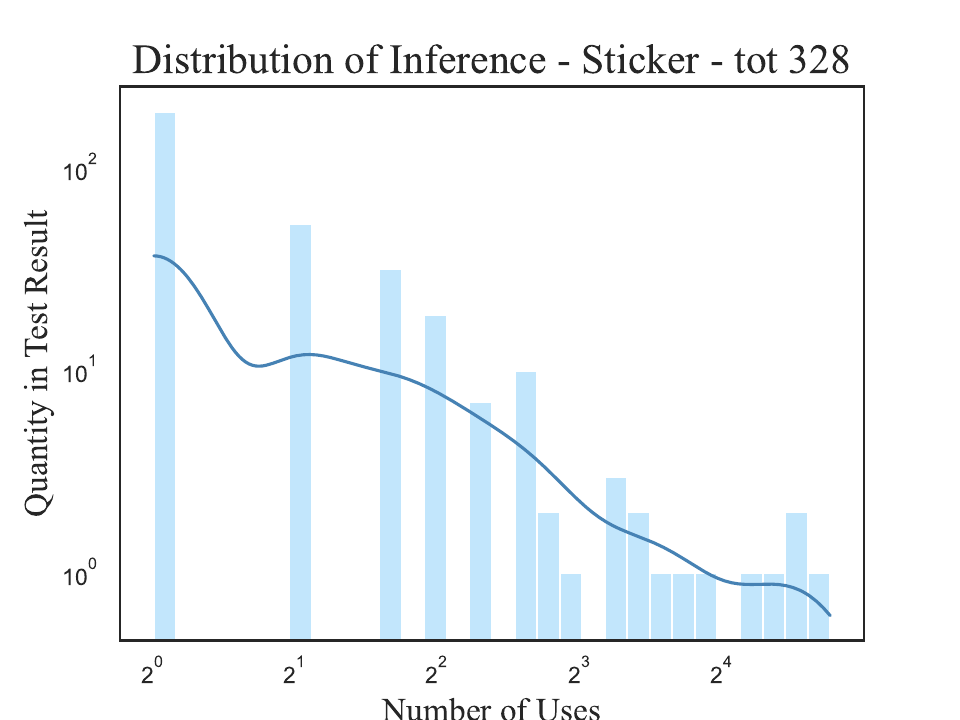}
          \subcaption{Distribution of stickers occurrences in test.}
          \label{fig:sticker_distribution_test}
        \end{minipage}\hfill
        \begin{minipage}{0.45\textwidth}
          \centering
          \includegraphics[width=\textwidth]{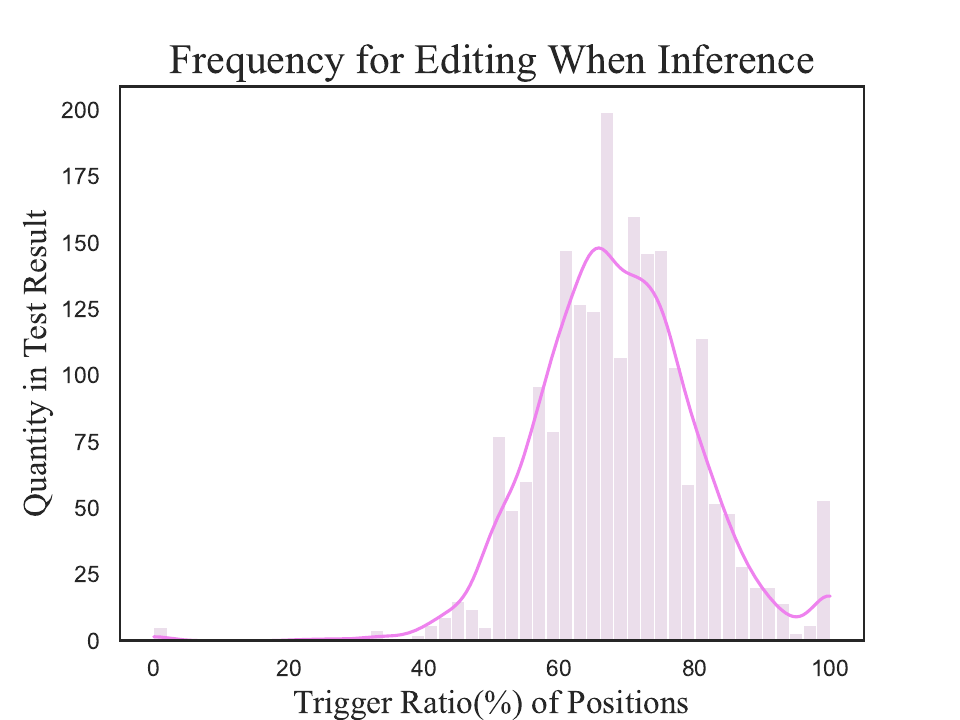}
          \subcaption{Distribution of trigger positions in test with prompting more trigger positions.}
          \label{fig:decoration_ratio_prompt}
        \end{minipage}\hfill
        \begin{minipage}{0.45\textwidth}
          \centering
          \includegraphics[width=\textwidth]{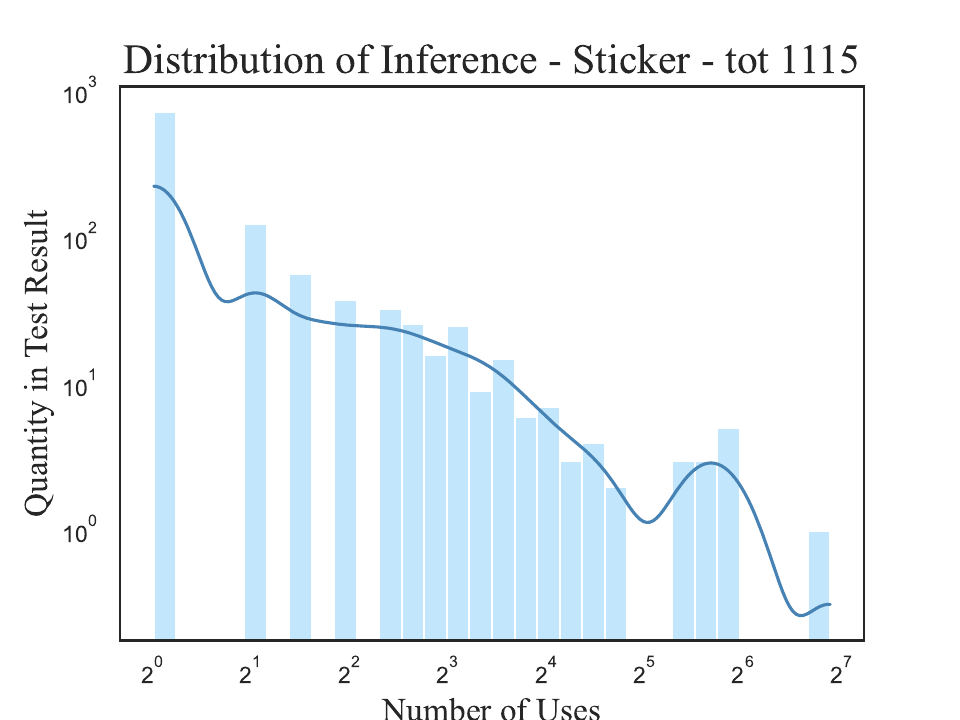}
          \subcaption{Distribution of stickers in test with prompting more stickers.}
          \label{fig:sticker_distribution_prompt}
        \end{minipage}
    \caption{Comparison of different prompt inference. Compared to the average of 50\% in Figure \ref{fig:decoration_ratio_test}, we observe a 70\% ratio of trigger positions when prompted with more trigger positions, as shown in Figure \ref{fig:decoration_ratio_prompt}. When prompted with more stickers, we obtain thousands of stickers, as illustrated in Figure \ref{fig:sticker_distribution_prompt}, whereas only 328 stickers are obtained without prompting, as seen in Figure \ref{fig:sticker_distribution_test}.
    }
    \end{figure}

    \subsection{Error Bar}
    \label{section:error bar}
    The standard error of the mean for the model in the last row of Table \ref{tab:main_result} is 0.54 for word precision, 0.46 for elem@word, 0.52 for elem@sentence, and 1.25 for the overall score. These values are obtained from five inferences conducted on two identically trained models.

\end{document}